\documentclass{article}

\usepackage[preprint,nonatbib]{neurips_2020}

\usepackage[utf8]{inputenc} 
\usepackage[T1]{fontenc}    
\usepackage{hyperref}       
\hypersetup{colorlinks,allcolors=black}
\usepackage{url}            
\usepackage{booktabs}       
\usepackage{nicefrac}       
\usepackage{microtype}      
\usepackage{subfiles}
\usepackage{xcolor}
\usepackage{multirow}
\usepackage{enumerate}
\usepackage{graphicx}
\usepackage[caption=false]{subfig}
\usepackage{amssymb, amsfonts, amsmath, dsfont}
\usepackage{makecell}
\usepackage{floatrow}
\usepackage[section]{placeins}
\usepackage[]{algorithm2e}

\DeclareRobustCommand{\Fig}[1]{Figure~\ref{fig:#1}}

\DeclareRobustCommand{\Eq}[1]{Equation~\ref{eq:#1}}

\def \E{\textrm{E}} 

\def \del{\partial}    

\def\bh{\textbf{h}}
\def\bx{\textbf{x}}

\def\by{\textbf{y}}

\def\bmu{\bvec{\mu}}

\def\be{\begin{equation}} 
\def\ee{\end{equation}} 
\newcommand \bea {\begin{eqnarray}} 
\newcommand \eea {\end{eqnarray}} 
 
\newcommand{\nn} {\nonumber}

\newcommand{\bvec}[1]{\boldsymbol{#1}}

\makeatletter
\newcommand*{\defeq}{\mathrel{\vcenter{\baselineskip0.5ex \lineskiplimit0pt
                     \hbox{\scriptsize.}\hbox{\scriptsize.}}}%
                     =}
\makeatother

\title{Neural Boltzmann Machines}

\author{
  Alex H. Lang\\
 Unlearn.AI\\
  \texttt{alex@unlearn.ai}\\
  \And
Anton D. Loukianov\\
  Unlearn.AI\\
  \texttt{aloukian@unlearn.ai}\\  
\And
Charles K. Fisher\\
  Unlearn.AI\\
  \texttt{drckf@unlearn.ai}\\ 
}

\begin{document}
\maketitle

\date{\today}

\begin{abstract}
Conditional generative models are capable of using contextual information as input to create new imaginative outputs. Conditional Restricted Boltzmann Machines (CRBMs) are one class of conditional generative models that have proven to be especially adept at modeling noisy discrete or continuous data, but the lack of expressivity in CRBMs have limited their widespread adoption. Here we introduce Neural Boltzmann Machines (NBMs) which generalize CRBMs by converting each of the CRBM parameters to their own neural networks that are allowed to be functions of the conditional inputs. NBMs are highly flexible conditional generative models that can be trained via stochastic gradient descent to approximately maximize the log-likelihood of the data. We demonstrate the utility of NBMs especially with normally distributed data which has historically caused problems for Gaussian-Bernoulli CRBMs. Code to reproduce our results can be found at \href{https://github.com/unlearnai/neural-boltzmann-machines}{https://github.com/unlearnai/neural-boltzmann-machines}.

\end{abstract}

\section{Introduction}

Conditional generative neural networks in a few years have gone from a side research interest of machine learning to the core technology in a product that has taken the world by storm\cite{openai2023gpt4}. This family of probabilistic models allow one to use contextual information or prompt $\bx$ to conditionally sample a new output $\by$ from the distribution $p(\by | \bx)$. It is now possible to use prompts to generate extremely high quality novel outputs in a variety of modalities, for example a text prompt can generate text answers\cite{openai2023gpt4}, images\cite{rombach2022highresolution}, or videos\cite{saharia2022photorealistic}. These advances have been driven by novel representations\cite{vaswani2017attention} that can be made deeper than previously thought possible\cite{hoffmann2022training} and combined with new conditional training techniques such as diffusion\cite{sohldickstein2015deep}.

However, there is another path forward which we introduce in this paper. Restricted Boltzmann Machines (RBMs)\cite{ackley1985learning} are a powerful class of energy based neural networks that model the world by learning a mapping between visible units that interact with data and hidden units which are an indirectly accessible latent space. RBMs have proven capable of learning complex distributions of discrete, categorical, or continuous data. RBMs have traditionally been used on unsupervised learning problems, but can be extended to a conditional generative model. The resulting networks are Conditional Restricted Boltzmann Machines (CRBMs)\cite{taylor2007modeling} and they have found applications in structure output prediction\cite{mnih2012conditional}, modeling motion style\cite{taylor2009factored}, and disease progression prediction\cite{fisher2019machine}.

RBMs and CRBMs played key roles at the start of the deep learning revolution\cite{ghojogh2022restricted}, but they have struggled to keep up with other model architectures that are capable of scaling to billions of parameters and taking advantage of double descent\cite{nakkiran2019deep}. The power of RBMs and CRBMs rests in the weights that couple visible and hidden units.  These weights are densely connected and can represent complex topics, but require iterative sampling to access their knowledge. Previous attempts have been made to make RBMs deeper including Deep Belief Networks (DBNs)\cite{hinton2006fast} and Deep Boltzmann Machines (DBMs)\cite{salakhutdinov2009deep} which required complex multistage training or approximately learned dynamics respectively. However, both DBNs and DBMs have fallen out of favor to regular RBMs, and instead state of the art RBMs achieve their best results by using a massive latent space that can be 10 times larger than the visible space\cite{liao2022gaussianbernoulli}.

In this paper we present Neural Boltzmann Machines (NBMs) which are a new extension of CRBMs that allow CRBMs to take full advantage of the push to deeper networks. We accomplish this by allowing all parameters of a CRBM to be their own neural network that is a function of the conditional input. This flexibility allows us to use arbitrary neural networks to represent the bias and variance of the visible units and the weights between visible and hidden units. We will demonstrate that NBMs are capable of using this new flexibility to tackle problems that were previously tricky for CRBMs to model correctly.

\section{Neural Boltzmann Machines}

A Neural Boltzmann Machine (NBM) is a conditional energy-based generative model that describes the distribution of $\by$ conditioned on $\bx$ by marginalizing over a vector of hidden units $\bh$. The joint distribution of the visible units $\by$ and the hidden units $\bh$ conditioned on the contextual information $\bx$ is
\be
p(\by, \bh | \bx) = Z^{-1}(\bx) e^{-U(\by, \bh | \bx)}
\ee
where the energy function is
\begin{align}
U(\by, \bh | \bx) 
= \frac{1}{2} (\by - \mu_{\theta}(\bx))' P_{\phi}(\bx) (\by - \mu_{\theta}(\bx))
-(\by - \mu_{\theta}(\bx))' W_{\psi}(\bx) \bh
\label{eq:nbm-energy}
\end{align}
and the normalizing constant is $Z(\bx) = \int d \by \sum_{H} e^{-U(\by, \bh | \bx)}$. In this paper we focus on the case in which the hidden units are discrete Ising-spins with $h_i = \pm 1$ (i.e., symmetrical Bernoulli variables), but one could use other types of hidden units in principle.

The energy function defined in \Eq{nbm-energy} is similar to the energy of a normal Restricted Boltzmann Machine (RBM) with Gaussian visible units, except the mean vector $\bmu_{\theta}(\bx)$, the diagonal precision matrix $P_{\phi}(\bx)$, and the weight matrix $W_{\psi}(\bx)$ matrix are functions of the context $\bx$ parameterized by $\theta$, $\phi$, and $\psi$, respectively. That is, we train feed-forward neural networks to output the parameters of an RBM. 

The purpose of the hidden units $\bh$ is to create a latent space to model the data distribution of interest, but the final goal is to sample the conditional distribution $p(\by | \bx)$ which can be obtained from $U(\by, \bh | \bx)$ by marginalizing over the hidden units via $p(\by | \bx) = \sum_{H} p(\by, \bh | \bx)$. The resulting marginal energy function or free energy is
\begin{align}
\mathcal{U}(\by | \bx) 
&=\frac{1}{2} (\by - \mu_{\theta}(\bx))' P_{\phi}(\bx) (\by - \mu_{\theta}(\bx)) 
- \log \sum_{H} e^{(\by - \mu_{\theta}(\bx))' W_{\psi}(\bx) \bh} \nn \\
&= \frac{1}{2} (\by - \mu_{\theta}(\bx))' P_{\phi}(\bx) (\by - \mu_{\theta}(\bx)) 
- {\bf 1}' \log \cosh \left( W_{\psi}(\bx)' (\by - \mu_{\theta}(\bx)) \right)
\label{eq:free-energy}
\end{align}

\subsection{Sampling NBMs}
As with a normal CRBM, we can use block Gibbs sampling to construct a Markov Chain that converges to $p(\by, \bh | \bx)$ by iteratively sampling $\bh | \by, \bx$ and $\by | \bh, \bx$, and then simply discarding the sampled hidden units.

The hidden units conditional distribution is
\be
\bh | \by, \bx \sim \text{Ising}(W_{\psi}(\bx)'(\by - \mu_{\theta}(\bx)) )
\ee
while the visible units conditional distribution for Gaussian visible units is
\be
\by | \bh, \bx \sim \mathcal{N}(\mu_{\theta}(\bx) + P_{\phi}(\bx)^{-1} W_{\psi}(\bx) \bh, P_{\phi}(\bx)^{-1})
\ee
or instead for discrete Ising-spins the visible conditional is
\be
\by | \bh, \bx \sim \text{Ising}(P_{\phi}(\bx) \mu_{\theta}(\bx) + W_{\psi}(\bx) \bh)
\ee

RBMs are prone to slow mixing in block Gibbs sampling, so additional methods such as Persistent Contrastive Divergence\cite{tieleman2008training} or Gibbs-Langevin sampling\cite{liao2022gaussianbernoulli} were invented to speed up Gibbs sampling mixing. However, we have not observed this mixing problem with NBMs. A key advantage of the parameterization of NBMs is that the Gibbs sampling starts from the outputs of a feedforward network that represents the bias and precision of the data. We hypothesize that mixing is less of a problem in general for conditional generative models because the entropy of $p(\by | \bx)$ is typically much lower than the entropy of $p(\by)$, and NBMs narrow the sampling search space even further by learning an explicit bias and precision.

\subsection{Training NBMs}

To train a NBM, we need to compute the gradients of the negative log-likelihood
\be
\mathcal{L}(\theta, \phi, \psi) \defeq - \E_{data}[\log p(\by | \bx)] \, .
\ee
Noting that $p(\by | \bx) = \mathcal{Z}(\bx)^{-1} e^{-\mathcal{U}(\by|\bx)}$, where $\mathcal{Z}(\bx) = \int d\by e^{-\mathcal{U}(\by|\bx)}$, we can compute the gradient with respect to the model parameters (e.g., $\theta$) using,
\begin{align}
\frac{\del \mathcal{L}}{\del \theta} 
&= - \frac{\del}{\del \theta} \E_{data}[\log p(\by | \bx)] \nn \\
&= \E_{data}[\frac{\del}{\del \theta} \mathcal{U}(\by|\bx)]
+ \E_{data}[\frac{\del}{\del \theta} \log \mathcal{Z}(\bx)] \nn \\
&= \E_{data}[\frac{\del}{\del \theta} \mathcal{U}(\by|\bx)]
+ \E_{data}[\frac{\int d\by  \frac{\del}{\del \theta} e^{-\mathcal{U}(\by|\bx)}}{\int d\by e^{-\mathcal{U}(\by|\bx)}}] \nn \\
&= \E_{data}[\frac{\del}{\del \theta} \mathcal{U}(\by|\bx)]
- \E_{data}[ E_{p(\by| \bx)}[\frac{\del}{\del \theta} \mathcal{U}(\by|\bx)]] .
\end{align}
The gradients with respect to $\phi$ and $\psi$ are similar. The first term (called the positive phase) involves taking the gradient of energy function and averaging over observed $(\bx, \by)$ pairs. The second term involves taking the gradient of the energy function and averaging over observed $\bx$ and generated $\by | \bx$. The gradient of the negative log-likelihood is just the difference of these two terms.

We can use standard backpropagation to compute gradients as long as $\mu_{\theta}(\bx)$, $P_{\phi}(\bx)$, and $W_{\psi}(\bx)$ are differentiable functions of the parameters. We note that to ensure $P_{\phi}(\bx)$ remains non-negative during training we perform a reparametization to learn $\log P_{\phi}(\bx)$ instead. Training then consists of the following steps. First, we sample a minibatch of data, $(\bx_i, \by_i)$, and perform a backwards pass through \Eq{free-energy}. Next, we perform $k$-steps of block Gibbs sampling to generate $\tilde{\by}_i$ conditioned on $\bx_i$ and perform a second backwards pass through \Eq{free-energy}. Finally, we compute the gradient of the negative log-likelihood by taking the difference between the two backwards passes, and update the model parameters with our chosen optimizer.  

\section{Experiments}

\subsection{Tips and Tricks}

Gaussian-Bernoulli RBMs are notoriously tricky to train and many machine learners have shed tears trying to get them to work\cite{liao2022gaussianbernoulli}. Researchers have proposed a variety of fixes including parallel tempering sampling\cite{6706831}, changes to the training algorithm\cite{10.1371/journal.pone.0171015}, or replacements to contrastive divergence\cite{9405414}.

In theory, Neural Boltzmann Machines could face some of the similar limitations, but we have found that in practice Neural Boltzmann Machines work out of the box with minimal tricks. There are two things we that have found to be crucial when working with NBMs. 

First, we normalize the outputs of the weight network $W_{\psi}(\bx)$ by $\sqrt{n_y}^{-1}$. The motivation for this is to better condition the matrix $P_{\phi}(\bx) - W_{\psi}(\bx) W_{\psi}(\bx)'$. When $\by$ is continuous, we can make a Laplace approximation around the point $\by = \mu_{\theta}(\bx)$, and the result of the math is that $\by = \mu_{\theta}(\bx)$ is a local minimum of the energy function as long as $P_{\phi}(\bx) - W_{\psi}(\bx) W_{\psi}(\bx)'$ is positive definite. One could add an additional loss during training such as logarithm of the determinant of the matrix, or a penalty on the logarithm of its condition number which would serve as a soft constraint on the positive definiteness of $P_{\phi}(\bx) - W_{\psi}(\bx) W_{\psi}(\bx)'$. However, we found that these additional losses were unnecessary as long as we appropriately normalized the output of $W_{\psi}(\bx)$.

Second, we have found that it is important to apply $L2$ penalties on $\mu_{\theta}(\bx)$, $P_{\phi}(\bx)$, and $W_{\psi}(\bx)$ and that one should set the penalty on $W_{\psi}(\bx)$ to be larger than $\mu_{\theta}(\bx)$ and $P_{\phi}(\bx)$. We have found that using a $L2$ penalty of $1.0$ on $W_{\psi}(\bx)$ and $0.5$ on $\mu_{\theta}(\bx)$ and $P_{\phi}(\bx)$ works well in practice across a wide variety of problems. If one does not set a larger $L2$ penalty on $W_{\psi}(\bx)$, we have found that NBMs are prone to failling into the trap of not using $\mu_{\theta}(\bx)$ and $P_{\phi}(\bx)$ and instead make a poor attempt to only use $W_{\psi}(\bx)$ to model everything.

Beyond those two key points, we have found we have been able to discard all other RBM tricks, and importantly do not need to do anything more complicated for the loss and sampling than standard Contrastive Divergence\cite{hinton2006training}. The only key parameter we have found important to optimize is the learning rate and we have found that binary visible units requires a higher learning rate than Gaussian visible units.

\subsection{Implementation}
We present results of training NBMs on two datasets: MNIST and FashionMNIST. Both datasets consists of grayscale images that are 28 by 28 pixels and have 10 classes each. We processed the grayscale images to be centered at zero and added normally distributed noise with mean zero and variance $0.001$ to ensure that every pixel is not trivially set to a single value.

For all the experiments, we trained for $50$ epochs with a batch size of $128$, $32$ MCMC sampling steps of which we only take the final output, and a $L2$ penalty of $1.0$ on $W_{\psi}(\bx)$ and $0.5$ on $\mu_{\theta}(\bx)$ and $P_{\phi}(\bx)$. The $W_{\psi}(\bx)$ matrix has $64$ hidden dimensions. The $\mu_{\theta}(\bx)$ and $P_{\phi}(\bx)$ are both 2 layer linear feed forward networks an intermediate dimension of $32$ and a ReLU activation between layers. We found that $\mu_{\theta}(\bx)$ and $P_{\phi}(\bx)$ could be a single layer but the training converges faster when they are deeper.

There are only two changes needed to switch from Gaussian visible units to Ising visible units. First, we use a learning rate of $0.0005$ for Gaussian visible units and a learning rate of $0.01$ for Ising visible units. Second, for a Bernoulli distribution, bias and precision are trivially coupled. Therefore, we add a Tanh activation layer to the end of the bias network to force the NBM to utilize the bias and precision nets differently.

\begin{figure}[ht]
\caption{MNIST conditional samples. Each column consists of 4 samples conditionally generated for each digit.}
\includegraphics[width=12cm]{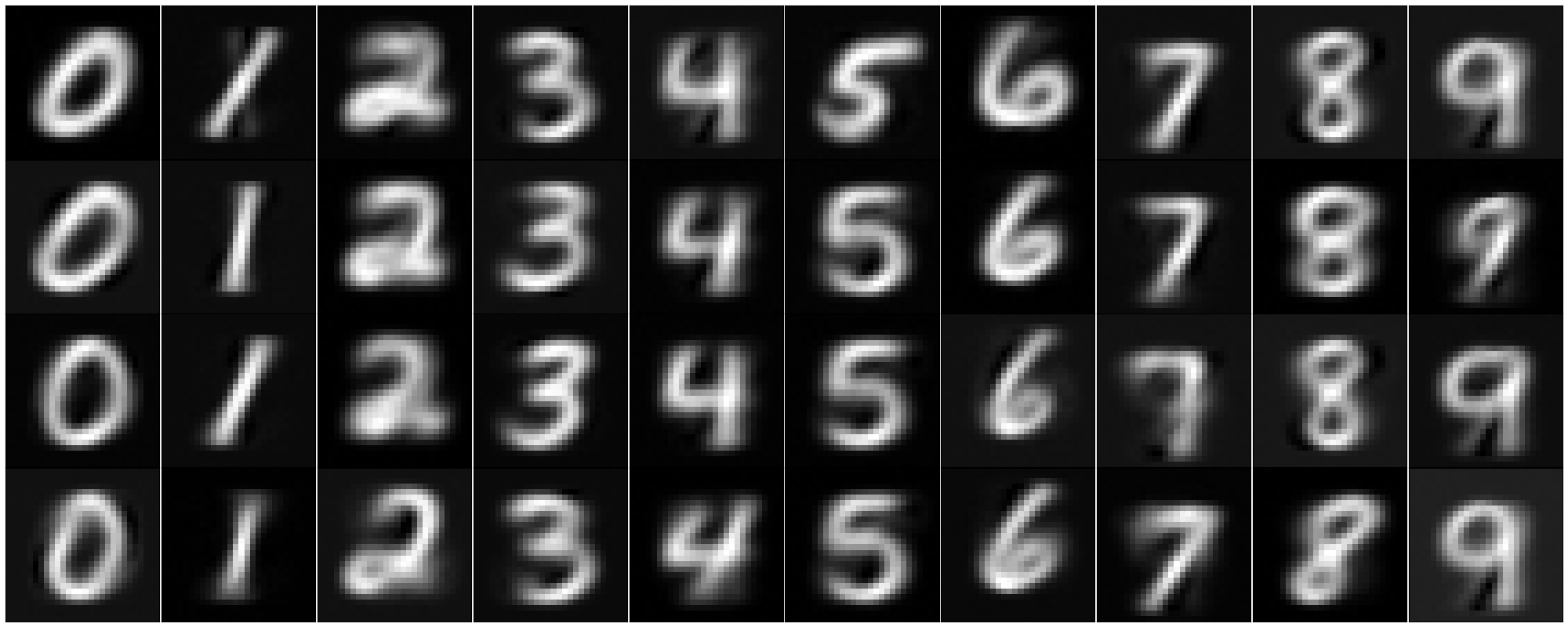}
\label{fig:mnist-samples}
\end{figure}

\begin{figure}[ht]
\caption{FashionMNIST conditional samples. Each column consists of 4 samples conditionally generated for each clothing item. The label of each column is T-shirt/Top (0), Pants (1), Pullover (2), Dress (3), Coat (4), Sandal (5), Shirt (6), Sneaker (7), Bag (8), and Ankle boot (9).}
\includegraphics[width=12cm]{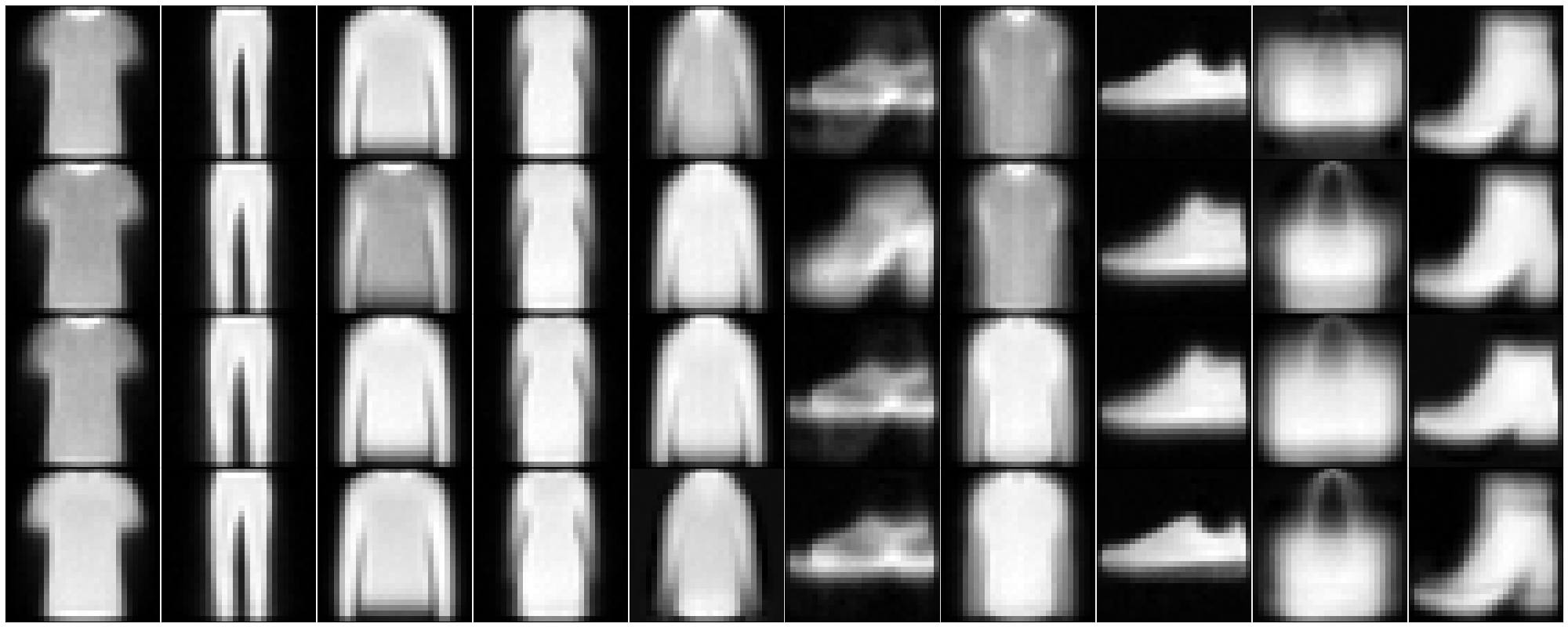}
\label{fig:fashion-mnist-samples}
\end{figure}

\subsection{MNIST Results}

\begin{figure}[ht]
\caption{MNIST conditional samples with NBM component outputs. Each column is conditioned on a different digit. The first row is the sampled output, the second row is the output from the bias network, the third row is the inverse output from the precision network, and the fourth row is the output from the weights network averaged over the hidden units.}
\includegraphics[width=12cm]{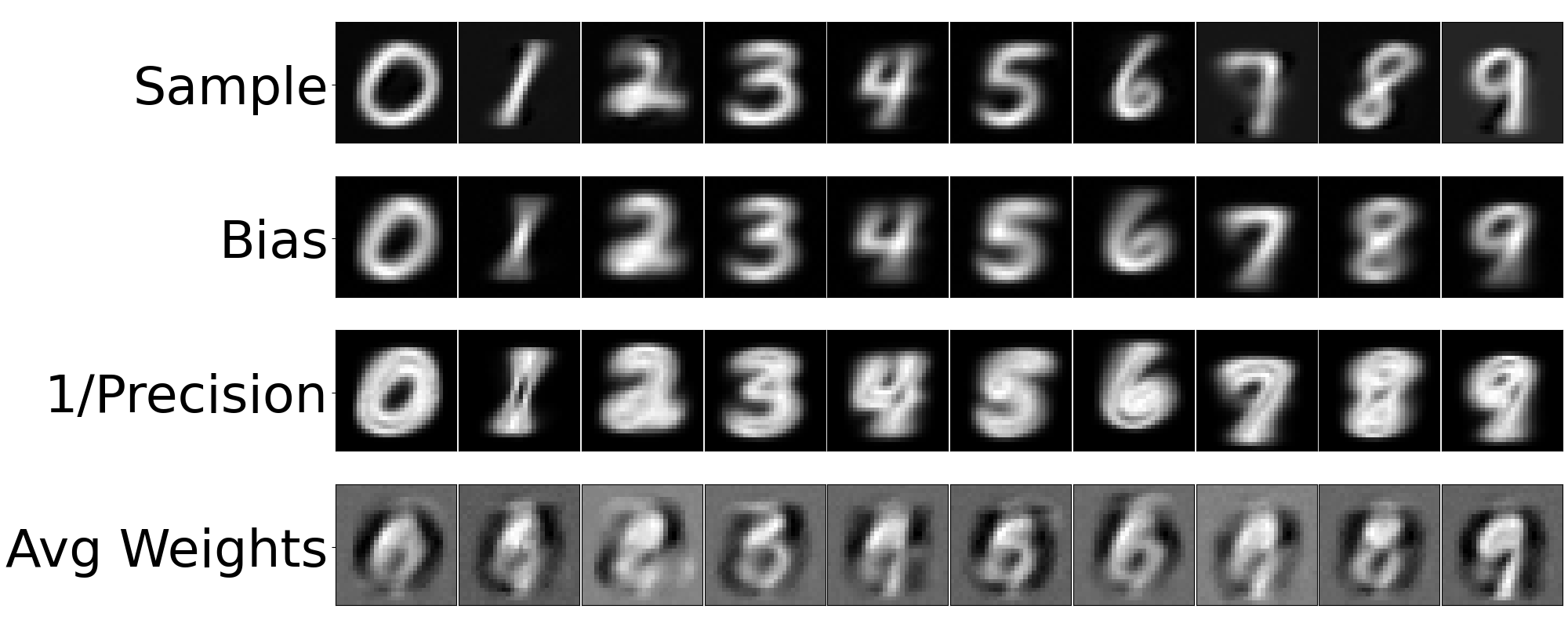}
\label{fig:mnist-details}
\end{figure}

Here we present results of training on Gaussian MNIST, but similar results can be reproduced with our code for Ising MNIST. In \Fig{mnist-samples}, we sample conditionally for each of the ten digits; taking the mean output of the last Gibbs sampling step. The resulting digits are sharp images which is often a struggle with Gaussian-Bernoulli RBMs. It is informative to inspect what each part of the NBM learns. \Fig{mnist-details} demonstrates that the NBM is able to learn the appropriate bias and precision (which we present as variance in order to have an easy comparison with bias). We can also look into the role of the weights by averaging over the hidden units and we see that the weights also learn representations of each digit that can be used to generate the full data distribution.

\subsection{FashionMNIST Results}

\begin{figure}[ht]
\caption{FashionMNIST conditional samples with NBM component outputs. Each column is conditioned on a different clothing item. The label of each column is T-shirt/Top (0), Pants (1), Pullover (2), Dress (3), Coat (4), Sandal (5), Shirt (6), Sneaker (7), Bag (8), and Ankle boot (9). The first row is the sampled output, the second row is the output from the bias network, the third row is the inverse output from the precision network, and the fourth row is the output from the weights network averaged over the hidden units.}
\includegraphics[width=12cm]{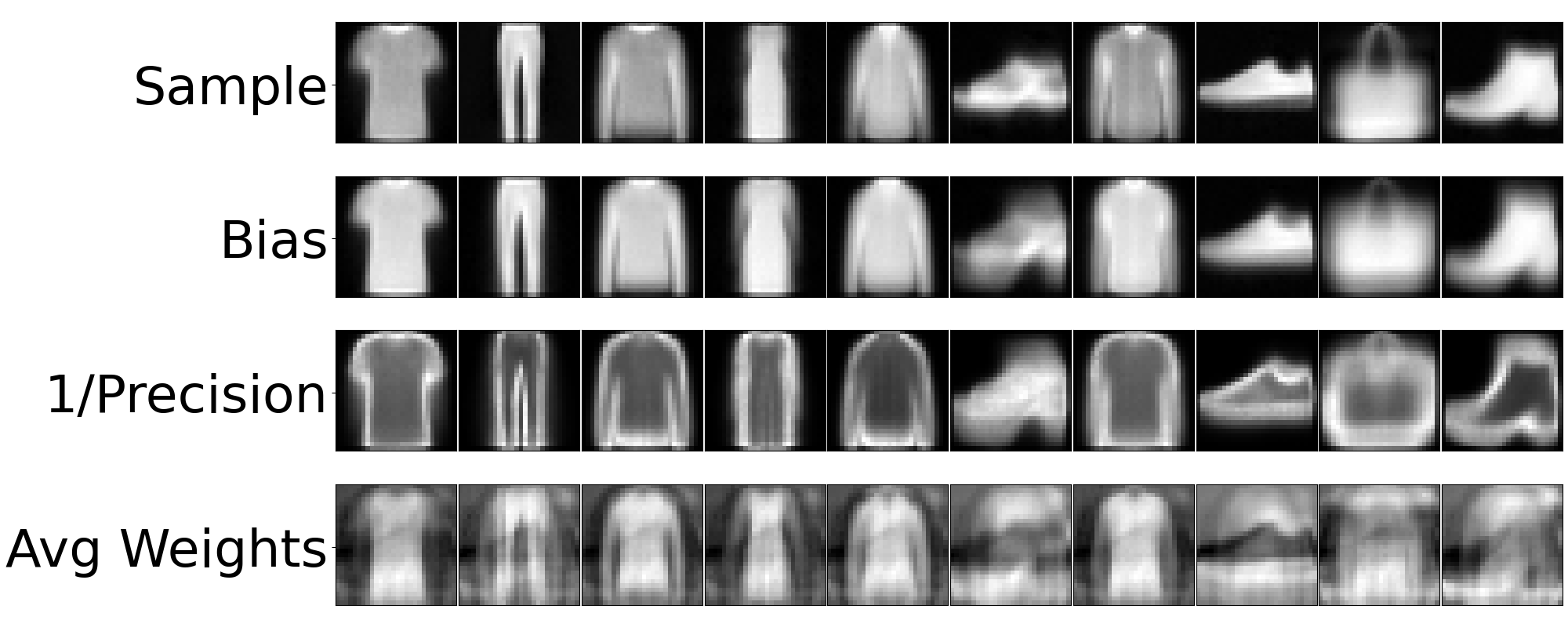}
\label{fig:fashion-mnist-details}
\end{figure}

While MNIST has served as a useful benchmark for machine learning, its simplicity limits its scope of generalizibility. We also present results on FashionMNIST which was designed to follow the same setup as MNIST to ensure comparability but made more challenging by having a wider distribution of data. In \Fig{fashion-mnist-samples} we see that the NBM is able to generate diverse samples and correctly models the two types of sandals as see in the sixth column. When we look into what the NBM has learned in \Fig{fashion-mnist-details}, we see that it has correctly used the bias to learn the core shape of an object and the precision to model the variation which is primarily at the edge of the objects. The weights capture the complexity of multiple modes. Similar to other RBMs, we see that the NBM struggles to capture fine grain textures in FashionMNIST but we made to attempts to tune the architecture of the parameters in the neural network components.

\section{Conclusion}

In this paper, we presented new class of conditional generative model called Neural Boltzmann Machines (NBMs). NBMs are extensions of RBMs and CRBMs that replace each parameter in a CRBM with its own feed forward network. This allows us to take full advantage of the deep learning revolution and allows us the flexibility to chain together arbitrary neural networks with a RBM. In essence, a typical feed forward network maps a point in $\bx$ to a point in $\by$, whereas an NBM maps a point in $\bx$ to an RBM that represents $p(\by | \bx)$. We demonstrate that NBMs that model both discrete and continuous variables and provide open source code at \href{https://github.com/unlearnai/neural-boltzmann-machines}{https://github.com/unlearnai/neural-boltzmann-machines} for others to extend our results. We look forward to novel ways to combine RBMs with the latest and greatest advances in machine learning.

\bibliographystyle{plain} 
\bibliography{refs} 

\end{document}